# Population Estimation using Deep Learning over Gandhinagar Urban Area

Jai Singla[a]* , Pearl Jotania[b], Keivalya Pandya[c]

[a] *Space Applications Centre[1], ISRO, Ahmedabad – 380015, India*

[b] *Charotar University of Science and Technology Changa, Gujarat*

[c] *Department of Mechanical Engineering, Birla Vishvakarma Mahavidyalaya, Vallabh Vidyanagar, India*

Corresponding Author : *jaisingla@gmail.com

## Abstract

Population estimation is crucial for various applications, from resource allocation to urban planning. Traditional methods such as surveys and censuses are expensive, time-consuming and also heavily dependent on human resources, requiring significant manpower for data collection and processing. In this study a deep learning solution is proposed to estimate population using high resolution (0.3 m) satellite imagery, Digital Elevation Models (DEM) of 0.5m resolution and vector boundaries. Proposed method combines Convolution Neural Network (CNN) architecture for classification task to classify buildings as residential and non-residential and Artificial Neural Network (ANN) architecture to estimate the population. Approx. 48k building footprints over Gandhinagar urban area are utilized containing both residential and non-residential, with residential categories further used for building-level population estimation. Experimental results on a large-scale dataset demonstrate the effectiveness of our model, achieving an impressive overall F1-score of 0.9936. The proposed system employs advanced geospatial analysis with high spatial resolution to estimate Gandhinagar's population at 278,954. By integrating real-time data updates, standardized metrics, and infrastructure planning capabilities, this automated approach addresses critical limitations of conventional census-based methodologies. The framework provides municipalities with a scalable and replicable tool for optimized resource management in rapidly urbanizing cities, showcasing the efficiency of AI-driven geospatial analytics in enhancing data-driven urban governance.

## Introduction

Precise population distribution estimation is essential for informing choices about emergency response, public service delivery, infrastructure planning, and city development. Traditional census-based approaches frequently fall behind the complexity and speed of urban change in fast-changing urban cities. Manual population estimation techniques find it difficult to capture the realities of today's urban landscapes as they become denser and more fragmented due to the emergence of residential zones, vertical expansion, and informal housing. These traditional methods usually produce low-resolution or out-of-date data, which reduces their ability to support accurate and timely urban management.

Cao and Weng (2024) developed a CNN based super resolution method to improve sentinel-1 and sentinel-2 satellite imagery to 2.5 meter for accurate building height estimation in the northern hemisphere. The study utilized a dataset of 45k samples collected from 301 global cities. Their CNN approach demonstrated superior performance compared to traditional methods, achieving RMSE values of 10.3m, 5.7m, 4.1m in china, USA and Europe countries respectively [1]. Metzger et al. (2024) combined SAR and optical imagery with dasymetric disaggregation to produce fine-scale population maps. The approach improved resolution over traditional census redistribution. Validation showed notable gains in urban areas with diverse surface materials [2]. Doda et al. (2024) introduced an explainable CNN for estimating urban population at ~1 km resolution. The model provided visual attribution maps to highlight key image regions driving predictions. Accuracy was competitive with black-box methods while improving interpretability [3]. Duan et al. (2024) demonstrated the use of a new Chinese VHR satellite sensor with CNNs for urban population mapping. The model produced detailed maps for metropolitan areas. Accuracy improved over Sentinel-based methods due to higher spatial resolution [4]. Chawla et al. (2024) proposed a semi-supervised framework for building height estimation using crowdsourced street-view imagery and OpenStreetMap footprints. In this approach, pseudo-labels generated from facade detections were used to train Random Forest, SVM, and CNN regressors. The model achieved a competitive Mean Absolute Error (MAE) of about 2.1 m in Heidelberg, Germany, while relying only on low-cost and open-source data. This demonstrates that semi-supervised fusion of image and morphometric features can effectively substitute for DEM or LiDAR in data-scarce regions, making the method particularly useful for large-scale urban population estimation [5]. Vergara (2024) presented a building-level population estimation model for Quezon City, Philippines, using Lidar-derived 3D building volume data combined with elevation models, building footprints, land use data, and local regulations. The study achieved a normalized absolute error (NAE) of 0.133 and a very high $R^2$ of approximately 0.976, indicating high predictive accuracy. However, the model displayed systematic biases—it tended to underestimate populations in high-density areas and overestimate in lower-density barangays. This work demonstrates how detailed 3D building information (volume, structure, land use) can produce highly accurate population estimates at micro-level scales, though it also highlights the need to address biases in diverse urban density contexts [6]. Swanwick et al. (2022) applied dsymmetric redistribution of census counts using impervious surface masks from Landsat. The dataset covered the entire US at 100 m resolution. The resulting maps achieved higher spatial accuracy than uniform redistribution [7]. Fibaek et al. (2022) trained CNN model on high-resolution satellite imagery and census data from multiple country to estimate population of different countries. the model addressed domain shift and transferability challenges. It achieved competitive accuracy across diverse regions but performance degraded when applied to geographically distinct areas. The study relying solely on 2D imagery [8]. Neal et al. (2022) explored using CNN-derived features from high-resolution imagery to predict population without census data. The model performed well in areas with consistent building morphology. However, performance dropped in heterogeneous urban environments. The study used only 2D spectral information [9]. Huang et al. (2021) evaluated several deep learning architectures (ResNet, VGG, Inception) for mapping population distribution from high-resolution imagery. The models used grid-cell level census data and neighborhood context features. Results showed improved accuracy when contextual spatial information was included [10]. Chen et al. (2021) developed a random forest regression approach to estimate population at building level by combining LiDAR data, POI data and nighttime light (NTL) data. The study was conducted on Shanghai city of China country. Their experimental results achieved coefficient of determination value 0.65 at building level [11]. Zong et al. (2019) proposed a hybrid deep learning approach combining SRCNN for spatial enhancement and LSTM for temporal modelling to produce high-resolution dynamic population maps. The model was trained on mobile phone data, census statistics, and POI datasets from Shanghai. It achieved superior spatial accuracy compared to traditional dasymetric mapping, improving temporal population variation capture. However, the method operated at grid-level [12].

Prior work has demonstrated both deep-learning-based gridded population mapping and dasymetric disaggregation methods (e.g., DeepDPM, WorldPop frameworks, and recent Sentinel-based DL approaches) as well as building-level estimators that fuse LiDAR, POI, and nighttime lights. However, most deep-learning population studies operate at grid scales or treat building height as an externally derived feature. To our knowledge, there is no prior end-to-end CNN-based study that jointly trains on very-high-resolution multispectral patches and co-registered DEM-derived building-height channels for building- level residential/non-residential classification and subsequent population estimation in an Indian city context, which motivates our work.

**Proposed Deep Learning Framework:**

To address these challenges, this study proposes a deep learning-based framework for automated, scalable, and accurate building classification using high-resolution satellite imagery (30 cm), Digital Elevation Models (DEMs) at 50cm grid intervals, and vector datasets. At the core of this framework lies a Convolutional Neural Network (CNN) utilizing both ReLU and LeakyReLU activation functions, enhanced by feature engineering techniques to eliminate correlated inputs and optimize computational efficiency.

The model is trained on approximately 48k labelled records containing both residential and non-residential, with residential categories further used for building-level population estimation using a sliding window technique, with validation through Google Earth imagery, achieving a high classification accuracy of 0.9986. The classified building data is stored as raster layers, which enables consistent and fine-grained population estimation across time and space.

## Dataset Detail

Gandhinagar, the capital of Gujarat, India, is a planned city with area of 326 km² and population of 2,08,299 according to census 2011 [12]. This city features a well-organized urban layout with 30 sectors, extensive green cover, and a diverse range of buildings, including residential complexes, commercial establishments, administrative offices, and institutional structures. WorldView-3 satellite target based imaging was used for this study. Stereo pair was acquired in Nov 2023. The image has a spatial resolution of 0.3 m and dimensions of $14{,}173 \times 16{,}892$ pixels. This offers very high-resolution coverage that is ideal for extracting building footprints. Authors generated a very high resolution DEM at 0.5m grid interval using the acquired stereo pair by employing the technique of satellite photogrammetry. The sub-meter resolution images are good for detailed building classification. Meanwhile, the sub meter-level coverage enables a wider context to be analysed. While Data distribution (Residential and Non-Residential) of labelled building footprints of Gandhinagar used for model training is represented in Fig 1. Green color depicts non-residential buildings whereas blue color stands for residential buildings in the Fig-1.

Table 1 Dataset Detail

| Satellite - Scene | Spatial resolution (meter) | Dimensions | Date of acquisition |
|---|---|---|---|
| Worldview 3 | 0.3 | $14173 \times 16892$ | Nov 2023 |
| DEM generated from Worldview 3 stereo pair | 0.5 | $14173 \times 16892$ | Mar 2024 |

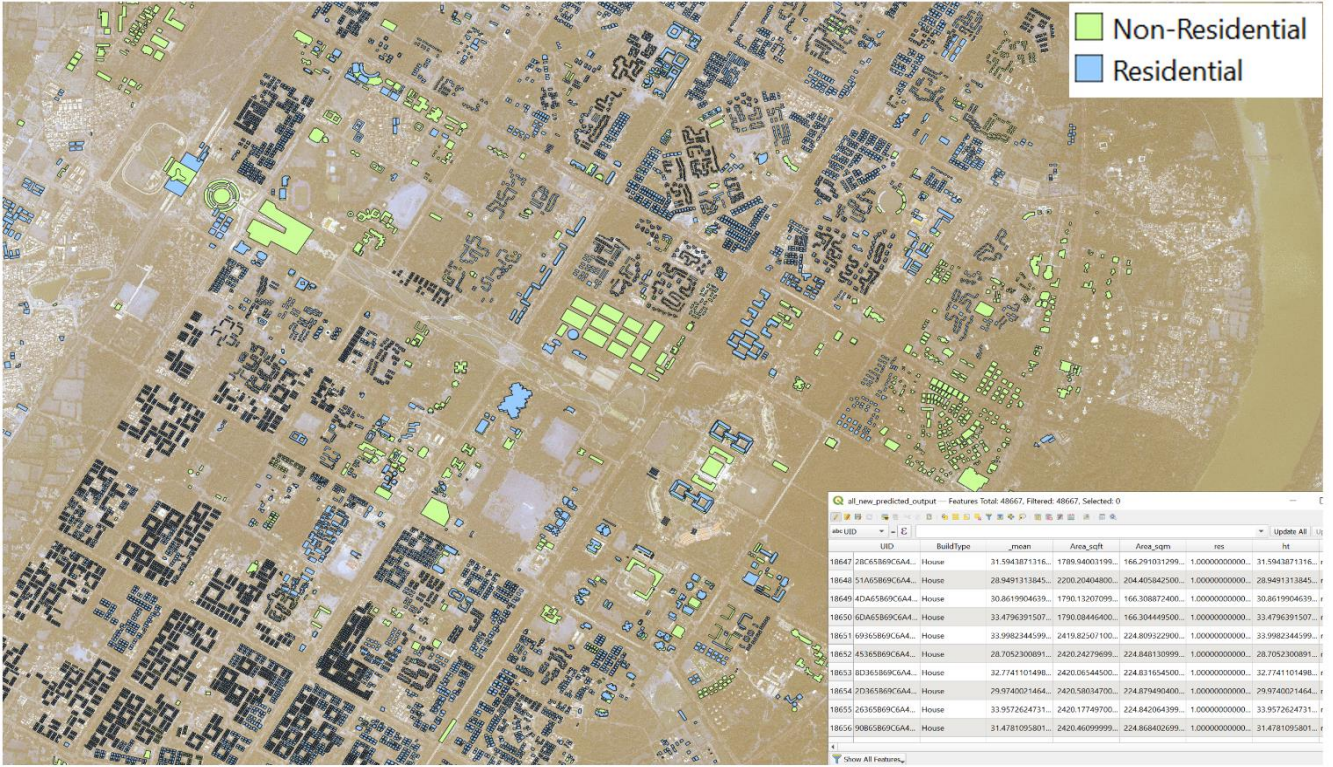

Figure 1 Data distribution of labelled building footprints of Gandhinagar used for model training

**Tools and Technologies**

Python 3.11 programming language [16] was used to train and validate proposed model and QGis [17] software was utilized to process raster and vector data.

**Data Pre-processing**

High resolution Satellite data, DEM data and vector files are utilized in this experiment to extract information of buildings and other features. We initially mapped a diverse range of building types from the dataset into two distinct groups, residential and non-residential, based on their characteristics. DEM and satellite imagery contains features like building height, area and shape of the building whereas unique identification number for each building (UID), building type (BuildType) and roof colour (RoofColor) are derived from vector data. We combined these features to generate derived attribute table. Derived attribute values are listed in Table 2.

Table 2 Derived attribute table from DEM , satellite images and vector data

|  | **Feature type** | **Mean** | **Max** | **Std** | **DType** |
|---|---|---|---|---|---|
| **UID** | Unique ID | - | - | - | object |
| **BuildType** | Categorical | - | - | - | object |
| **RoofColor** | Categorical | - | - | - | object |
| **_mean** | Continuous | 25.2176 | 78.7562 | 3.3335 | float64 |
| **Area_sqft** | Continuous | 1552.0230 | 191268.2879 | 4037.1331 | float64 |
| **Area_sqm** | Continuous | 144.1876 | 17769.4054 | 375.0619 | float64 |
| **res** | Categorical | - | - | - | float64 |
| **ht** | Continuous | 7.7176 | 61.2562 | 3.3335 | float64 |
| **geometry** | Polygon | - | - | - | Object (3) |

Mean, Max, and standard deviation (Std) values mentioned in table 2 are continuous variables representing statistical metrics of building characteristics. Building area in square feet and square meter are represented using Area_sqft and Area_sqm respectively. Building is residential or non-residential is represented in res variable. Building height in meter and footprint is denoted by ht and geometry respectively. Figure 2 and 3 represents correlation metrics before and after data pre-processing. By comparing this metrics, we can observe the changes in the correlation between the features.

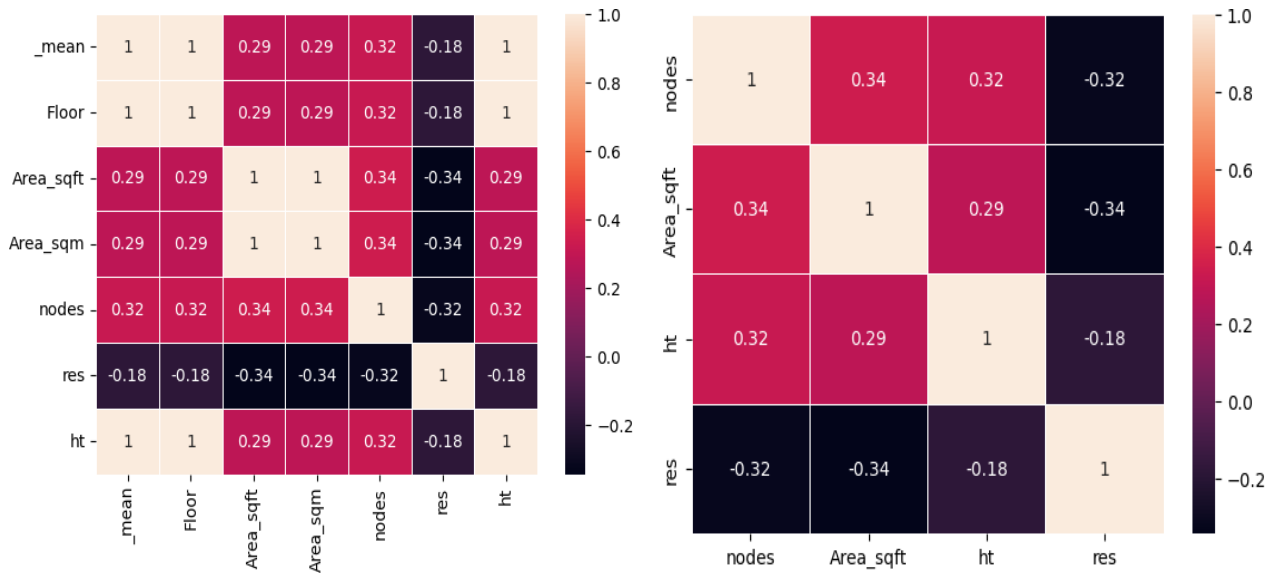

Figure 2 Correlation matrix of features of dataset   Figure 3 Correlation metrics of features after feature extraction

## Methodology

In this work, a dual-model framework was implemented, utilizing a CNN model to effectively classify building types into residential and non-residential categories, alongside ANN to precisely estimate population. Initially, Vector shapefile of the studied area was extracted using automated extraction techniques (Singla et al., 2023). This vector file was utilized as one of the input in this study. The proposed system level diagram is presented in Figure 4. An image patch with size 64 x 64 was extracted for each building by centering a window at the building's centroid within the satellite image. This window covers a ground footprint of approx. 32m x 32 m. Invalid or boundary-overlapping patches were discarded. Patches were normalized to the range [0,1] and stored as 4-band arrays. CNN model was trained on 15,999 labelled data which contains 417 (2.61%) non-residential buildings and 15,582 (97.3%) residential buildings. Dataset were split in training, testing and validation with 80:10:10 ratio respectively. Further this trained model is utilized to predict building type of rest of the 48k unlabeled data. This unlabeled data contains approximately 14k non-residential buildings like hospitals, universities and government buildings and approximately 34k residential buildings. Obtained predicted vector file is validated using Google Earth [18].

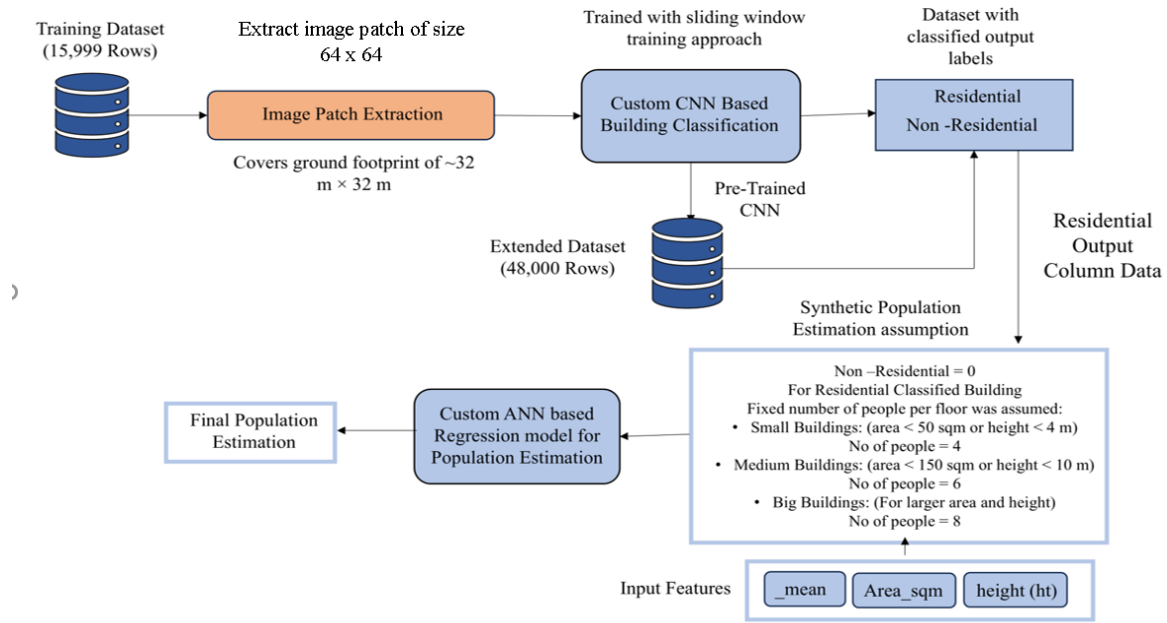

Figure 4 Workflow Diagram

**CNN Network Architecture**

A custom Convolutional Neural Network (CNN) was implemented to classify building patches as residential (class 1) or non-residential (class 0). As illustrated in Figure 5, image patches of size 64x64 with 4 spectral bands were utilized as inputs to the CNN model. The architecture consists three convolutional layers with 3x3 kernels, containing 32, 64 and 128 filters respectively, each followed by a ReLU activation function. After the convolutional layers, the output is flattened and passed to a fully connected layers for the classification task. Binary cross entropy and Adam optimization is employed with a learning rate of 1e-4, which helps in efficient and stable training of model. Hyper parameters used to train this CNN network is listed in Table 3. To address data imbalance and spatial diversity, a sliding window approach was employed. The labelled dataset was partitioned into overlapping windows (1k records per window), with each window further split into training and testing set in an 80:20 ratios. The CNN model was trained for 10 epochs on each window and the predicted labels for the test splits were stored in shape files. This approach improves the model's ability to work well in various urban areas and reduces the chances of overfitting.

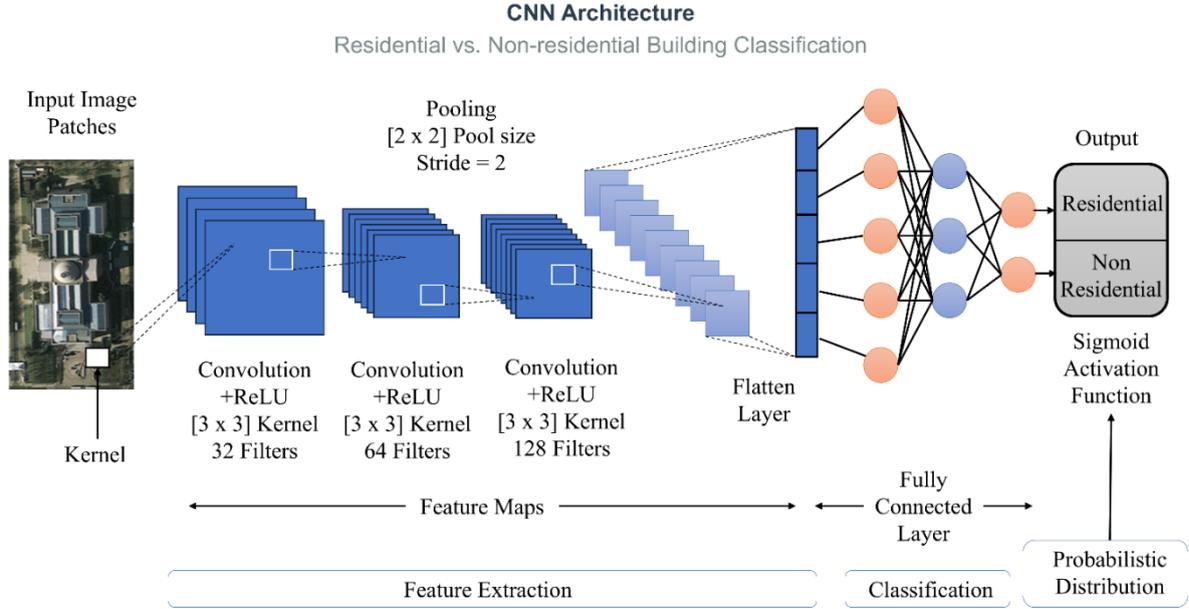

Figure 5 Convolution Neural Network Architecture

Table 3 Hyper parameters values for CNN model

| Input Image size | $64 \times 64 \times 3$ |
|---|---|
| Batch size | 32 |
| Train : validation | 80 : 20 |
| Learning rate | 0.0001 |
| Regularization | L2 = 1e-4 (dense layers) |
| Epoch | 10 |
| Drop rate | 0.5 |
| Loss Function | Binary cross entropy |
| Activation Function | ReLU (hidden layers), Sigmoid (output) |

Since building-level ground truth population data was unavailable, a synthetic labelling function was developed to generate population estimates based on interpretable, rule-based heuristics. These synthetic labels serve as training targets for the regression model.

To generate approximate ground truth labels for training the population regression model, a rule-based function was implemented using building characteristics. Buildings classified as non-residential were assigned a population of zero. For the remaining buildings population was estimated based on their area in square foot and height in meters. We assumed the number of residents per floor as 4 for small buildings (area < 50 or height < 4), 6 for medium-sized buildings (area < 150 or height < 10) and 10 for large buildings [19]. The assumption is done based on housing policy thresholds and urban dwelling characteristics. The number of floors was calculated using the formula:

$$\text{Floors} = \text{height} / 3 \text{ (assuming 3 meters per floor, minimum 1)} \qquad (1)$$

The estimated population for each building was determined by multiplying the number of floors by the corresponding number of residents per floor.

**ANN Model Architecture**

Instead of relying solely on rule-based calculations, this study introduces a second Artificial Neural Network (ANN) to directly predict the estimated population of residential buildings using a regression-based approach.

ANN model used to estimate population for this study is presented in Figure 6. This model builds upon the output of the binary classification stage (residential vs. non-residential). Residential buildings (res = 1) along with their synthetic population values were used to train the model, which performs population regression to predict the number of populations in each building. Input features for this model include _mean (mean pixel intensity of the image patch), Area_sqft (building footprint area) and ht (building height in meters) were normalized using standard scaling to improve model performance. The model architecture comprises three dense layers with 64, 32 and 8 respectively. Each of the layers uses the ReLU activation function to introduce non-linearity, enabling the model to learn complex patterns in the data. Output Dense layer uses a linear activation function to produce continues population value. The hyper parameters values used for training the model is discussed in table 4.

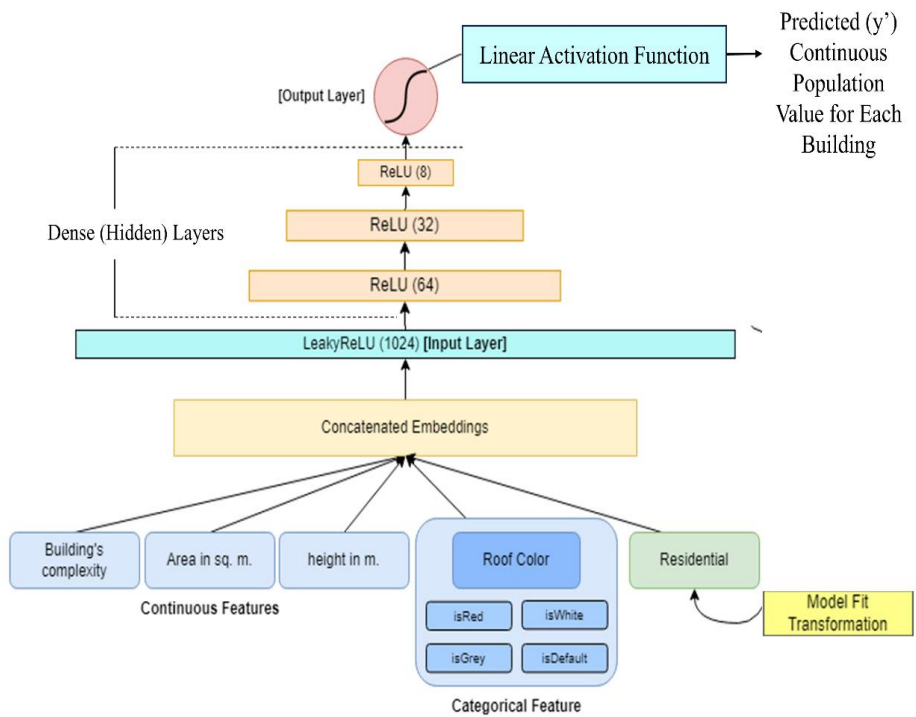

Figure 6 ANN Architecture for Population Estimation

Table 4 Hyper parameter value for ANN model

| Input Features | 3 (Area_sqm, ht, CNN classification output ) |
|---|---|
| Batch size | 64 |
| Train : validation | 80 : 20 |
| Learning rate | 0.0001 |
| Epoch | 20 |
| Drop rate | 0.3 |
| Loss Function | Mean Squared Error (MSE) |
| Activation Function | ReLU (hidden layers), Sigmoid (output) |
| Feature scaling | Min–Max normalization |

## Evaluation Metrics

Following accuracy metrics were used to evaluate the results of deep learning models:

**Accuracy: -** Accuracy is the ratio of correctly classified features to the total number of features. Accuracy is calculated using Equation 2.

$$\text{Accuracy} = \frac{TP + TN}{TP+TN+FP+FN} \qquad (2)$$

where TP=True Positive, FP=False Positive, FN=False Negative and TN=True Negative.

**Precision: -** Precision refers to the number of true positives (TP) divided by the total number of positive predictions. Precision can be calculated as Equation 3.

$$\text{Precision} = \frac{TP}{TP+FP} \qquad (3)$$

**Recall: -** Recall is the ratio of true positive to the sum of true positive and false positive as shown in Equation 4.

$$\text{Recall} = \frac{TP}{TP+FN} \qquad (4)$$

**F1 Score: -** F1 score combines precision and recall using harmonic mean. Equation 5 represents F1 Score.

$$\text{F1 Score} = \frac{2*\text{Precision}*\text{Recall}}{\text{Precision} + \text{Recall}} \qquad (5)$$

## Result and Discussion

The proposed CNN model demonstrated an excellent ability to classified buildings, labelling them residential and non-residential. From the confusion matrix shown in Figure 7, it can be observed that 78 out of 84 true non-residential buildings were properly classified, while 5 were misclassified as residential. Additionally, among the residential buildings, 3113 out of 3117 were correctly classified, while 4 were misclassified as residential.

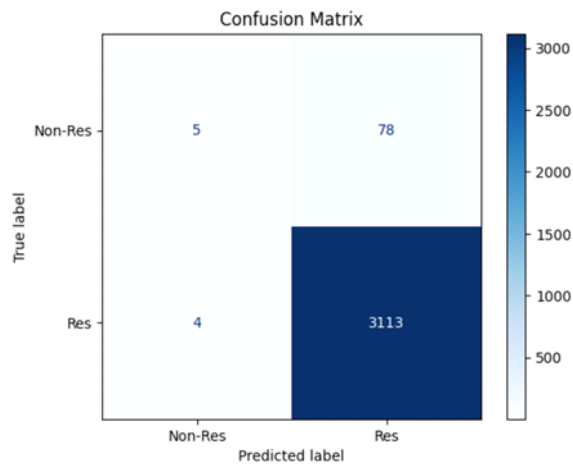

Figure 7 Confusion matrix

The prediction results of models are illustrated in Figure 11, 12, 13 and 14. In each figure, part (A) represents the predicted results of model and part (B) represent the validations on Google Earth. Correctly classified non-residential and residential buildings by the model are shown in Figure 11 and 12 respectively whereas residential building wrongly classified as non-residential building is shown in figure 13. Figure 14 shows the non-residential building wrongly classified as residential building.

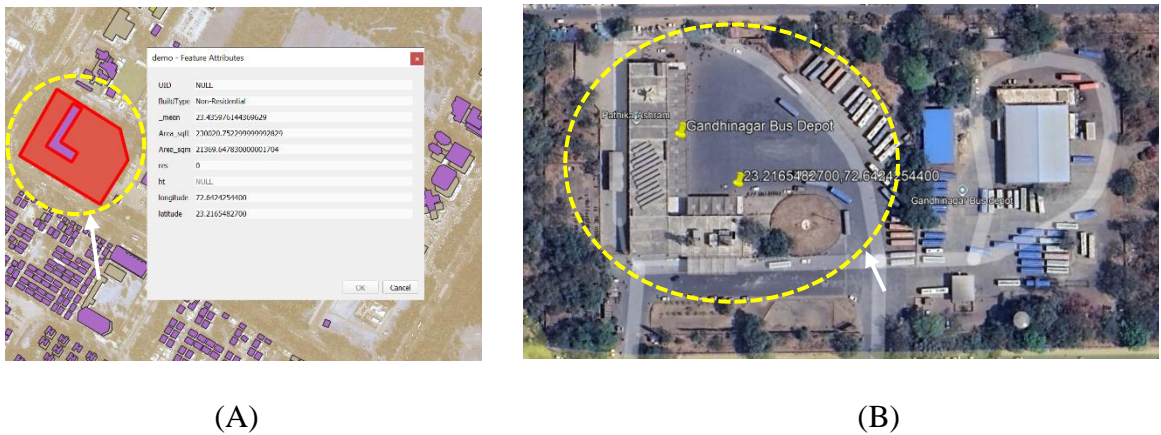

(A)                                      (B)

Figure 11. Actual Non-Residential and correctly classified as Non-residential

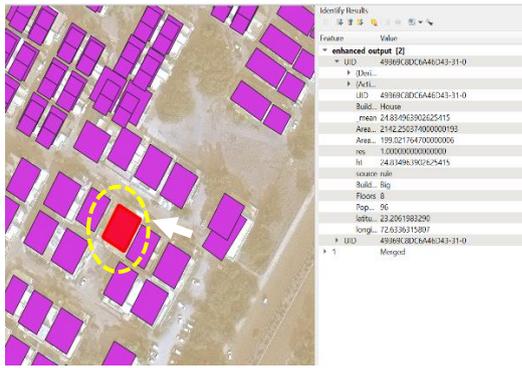 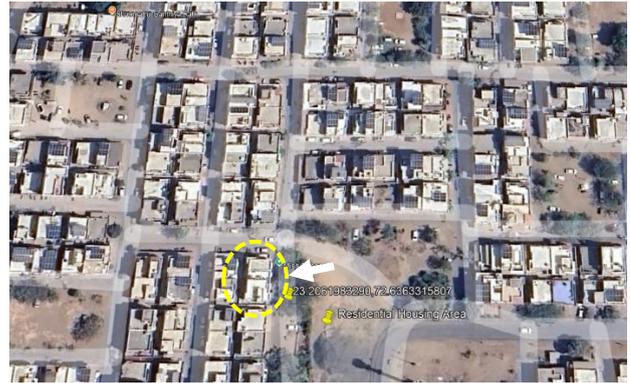

(A) (B)

Figure 12. Actual Residential and correctly classified as Residential

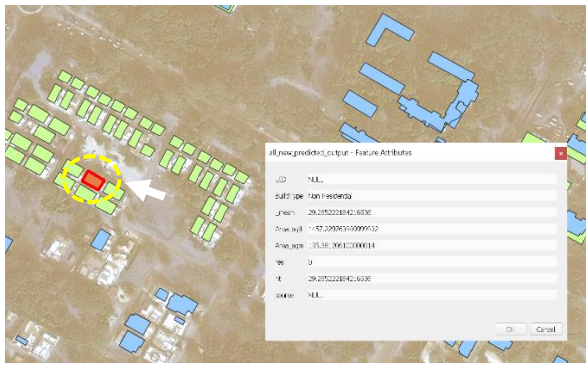 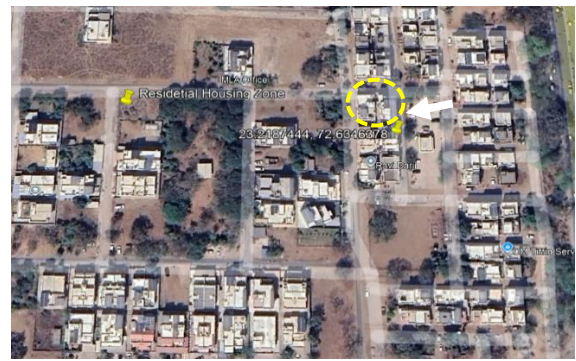

(A) (B)

Figure 13. Actual Residential and incorrectly classified as Non-Residential

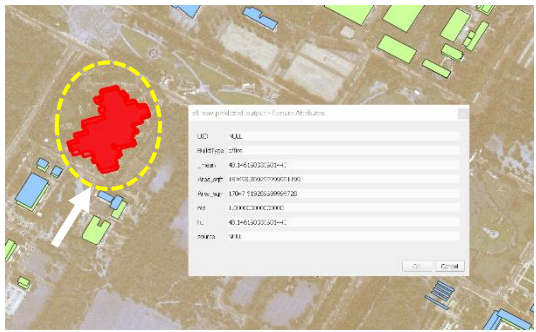 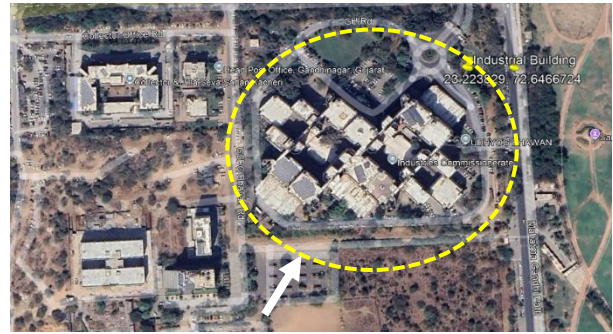

(A) (B)

Figure 14. Actual Non-Residential and incorrectly classified as Residential

The analysis of Table 5 indicates an average precision of 0.89, an average recall of 0.86 and an average F1 score of 0.88 achieved by proposed CNN model. The precision, Recall and F1 score of non-residential buildings are lower compare to residential buildings because it is difficult to distinguish between mixed-uses and institutional buildings due to their similar characteristics.

Table 5 Evaluation metrics values

|  | Precision | Recall | F1-Score |
|---|---|---|---|
| **Non-residential** | 0.79 | 0.73 | 0.76 |
| **Residential** | 0.99 | 0.99 | 0.99 |
| **Average** | 0.89 | 0.86 | 0.88 |

The AUC curve, shown in the figure 15, indicates the model's substantial capability to discriminate between non-residential and residential buildings. An area under the curve approaching 1 reflects excellent separation of the two classes. The model loss during regression training stage is represented Figure 16, which shows smooth convergence with no signs of overfitting.

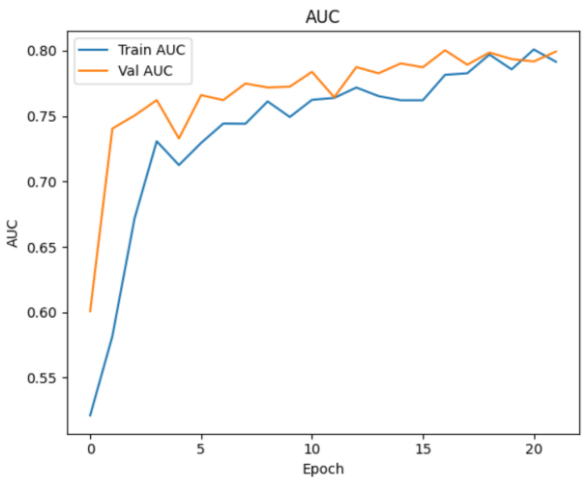

Figure 15 AUC curve

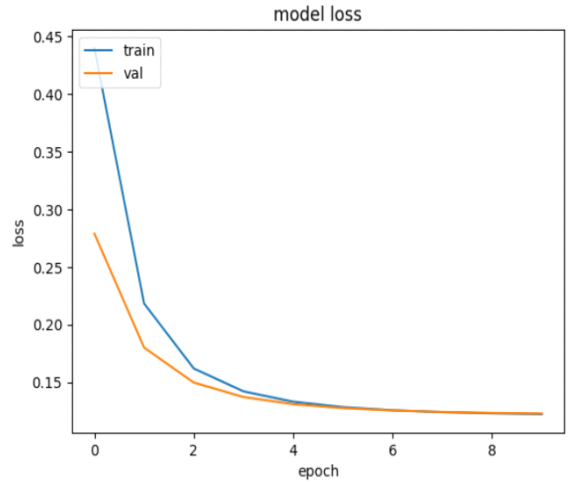

Figure 16 model loss curve

Population estimation for the urban area of Gandhinagar city using our proposed method is presented in Table 6, which corresponds with the census-projected figure of approx. 3 lakhs in 2025 [13]. These finding indicate that our model effectively captures population distribution. The consistency of the results with official projections and census data further substantiates the accuracy and reliability of the proposed deep learning approach.

Table 6 Population Estimation Results for Gandhinagar Urban Area

| Building Size | Number of Residential Buildings | Estimated Floors | Estimated Population |
|---|---|---|---|
| Small | 12,804 | 2–3 | 89,425 |
| Medium | 18,532 | 7–9 | 116,504 |
| Large | 2,358 | 10–15 | 68,957 |
| Total | 33,694 | - | 278,954 |

While the population estimate of 278,954 generated from this study is appropriate for general planning purposes, we cannot discount potential over or underestimations due to variability in occupancy or use of buildings. The official census done in 2011, lists Gandhinagar's population as 206,167 (metropolitan- 6,361,084) [13] [15], other demographic resources yield an aligned population number of 208,299 for the same year [14]. As it stands, the city's 2025 population is estimated to be around 300,000, based on previous growth levels [13] [15]. Future testing of this methodology, through use of current census information or physical ground surveys, will ultimately improve any veracity of the estimates obtained through this analysis. This base level of planning tool can provide urban planners with an effective, scale- able tool in which to examine building-level information in their strategic targeting of investments.

## Conclusion

This study proposes an automated population estimation workflow employing a dual architecture comprising CNN and ANN. The methodology achieved F1 score of 0.9936, accurately classifying 49k buildings in an urban area over Gandhinagar city and estimating population of 278, 954.Unlike traditional methods that depends on structures and tabular features, this approach uses high resolution satellite imagery to train the mode, allowing it to learn directly from the complex visual features such as Rooftop texture, shadows, green areas, parking lots etc. Some limitations include; possible bias during the training process of the dataset derived from class imbalance, the research needs to be verified in other urban areas, and the estimates about population were uniform occupancy rates with varied occupancies between buildings depending on mixed commercial activity in areas identified as residential. Conducting ground surveys will also help to enhance the accuracy of the results. However, this research needs to be verified in other urban areas to validate its reliability.